\documentclass[11pt,a4paper]{article}
\usepackage[hyperref]{emnlp2020}
\usepackage{times}
\usepackage{latexsym}

\usepackage{xspace}
\usepackage{arydshln}
\usepackage{multirow}
\usepackage{graphicx}  
\usepackage{CJKutf8}
\usepackage[ruled,linesnumbered,commentsnumbered]{algorithm2e}
\usepackage{subfig}
\usepackage{graphics}
\usepackage{balance}
\usepackage{float}
\usepackage{color}
\usepackage{CJK}

\usepackage{amsmath}
\usepackage{enumitem}
\usepackage{amsfonts}
\usepackage{multicol}
\usepackage{CJKutf8}

\usepackage{microtype}

\aclfinalcopy 

\title{On the Sparsity of Neural Machine Translation Models}

\author{Yong Wang\thanks{~~Work was done when interning at Tencent AI Lab.} \\ The University of Hong Kong \\ {\tt wangyong@eee.hku.hk}   \And
Longyue Wang \\ Tencent AI Lab \\ {\tt vinnylywang@tencent.com} \AND
Victor O.K. Li \\ The University of Hong Kong \\ {\tt vli@eee.hku.hk} \And
Zhaopeng Tu \\ Tencent AI Lab \\ {\tt zptu@tencent.com}
}

\date{}

\begin{document}
\maketitle
\begin{abstract}
Modern neural machine translation (NMT) models employ a large number of parameters, which leads to serious over-parameterization and typically causes the underutilization of computational resources. In response to this problem, we empirically investigate whether the redundant parameters can be reused to achieve better performance. Experiments and analyses are systematically conducted on different datasets and NMT architectures. We show that: 1) the pruned parameters can be rejuvenated to improve the baseline model by up to +0.8 BLEU points; 2) the rejuvenated parameters are reallocated to enhance the ability of modeling low-level lexical information.

\end{abstract}

\section{Introduction}
Modern neural machine translation~(NMT)~\cite{bahdanau2014neural, gehring2017convolutional,vaswani2017attention} models employ sufficient capacity to fit the massive data well by utilizing a large number of parameters, and suffer from the widely recognized issue, namely, over-parameterization. For example, \newcite{see2016compression} showed that over 40\% of the parameters in an RNN-based NMT model can be pruned with negligible performance loss. However, the low utilization efficiency of parameters results in a waste of computational resources \cite{qiao2018neural}, as well as renders the model stuck in a local optimum~\cite{han2017dsd,yu2019layer}. 

In response to the over-parameterization issue, network pruning has been widely investigated for both computer vision (CV) \cite{han2016deep,luo2017thinet} and natural language processing (NLP) tasks \cite{see2016compression,lan2019albert}. Recent work has proven that such spare parameters can be reused to maximize the utilization of models in CV tasks such as image classification~\cite{han2017dsd,qiao2018neural}. The leverage of parameter rejuvenation in sequence-to-sequence learning, however, has received relatively little attention from the research community. In this paper, we empirically study the efficiency issue for NMT models.

Specifically, we first investigate the effects of weight pruning on advanced Transformer models, showing that 20\% parameters can be directly pruned, and by continuously training the sparse networks, we can prune 50\% with no performance loss. Starting from this observation, we then exploit whether these redundant parameters are able to be re-utilized for improving the performance of NMT models. Experiments are systematically conducted on different datasets (i.e. Zh$\Rightarrow$En, De$\Rightarrow$En and En$\Rightarrow$Fr) and NMT architectures (i.e. Transformer, RNNSearch and LightConv). Results demonstrate that the rejuvenation approach can significantly and consistently improve the translation quality by up to +0.8 BLEU points. Further analyses reveal that the rejuvenated parameters are reallocated to enhance the ability to model the source-side low-level information, lacking of which leads to a number of problems in NMT models~\cite{tu2016modeling,dou2018exploiting,emelin2019widening}.

\paragraph{Contributions}
Our key contributions are:
\begin{itemize}
\setlength\itemsep{-0.1em}
\item We try early attempts to empirically investigate parameter rejuvenation for NMT models across different datasets and architectures.

\item We explore to interpret where the gains come from in two perspectives: learning dynamics and linguistic insights.
\end{itemize}

\section{Approach}
A standard NMT model directly optimizes the conditional probability of a target sentence $\mathbf{y} = y_1, \dots, y_{J}$ given its corresponding source sentence $\mathbf{x} = x_1, \dots, x_{I}$, namely $P(\mathbf{y}|\mathbf{x}; \theta) = \prod_{j=1}^{J} P(y_j | \mathbf{y}_{<j}, \mathbf{x}; \theta)$,
where $\theta$ is a set of model parameters and $\mathbf{y}_{<j}$ denotes the partial translation. The parameters of the NMT model are trained to maximize the likelihood of a set of training examples. Given a well-trained NMT model, we first prune its inactive parameters, and then rejuvenate them. Our implementation details are as follows. 

\paragraph{Pruning}
The redundant parameters in neural networks can be pruned according to a certain criterion while the left ones are significant to preserve the accuracy of the model. Specifically, we mask weight connections with low magnitudes in the forward pass and these weights are not updated during optimization. Given the weight matrix $W$ with $N$ parameters, we rank the parameters according to their absolute values. Supposed that the pruning ratio is $\gamma$ (i.e. $\gamma$\% of parameters should be pruned), we keep the top $n$ parameters ($n = N \times (1-\gamma)$), and remove the others with a binary mask matrix, which is the same size of $W$. We denote the pruned parameters as $\theta_p$, subject to $\theta_p\subset\theta$. There are two pruning strategies~\cite{liu2018rethinking}: 1) local pruning, which prunes $\gamma$\% of parameters in each layer; and 2) global pruning, which compares the importance of parameters across layers. Following \citet{see2016compression}, we retrain the pruned networks after the pruning phase. Specifically, we continue to train the remaining parameters, but maintain the sparse structure, that is we optimize $P(\mathbf{y}|\mathbf{x}; \theta)$ with the constraint: $a=0, \forall a\in\theta_p$.

\begin{figure}[t]
\centering
\includegraphics[width=0.42\textwidth]{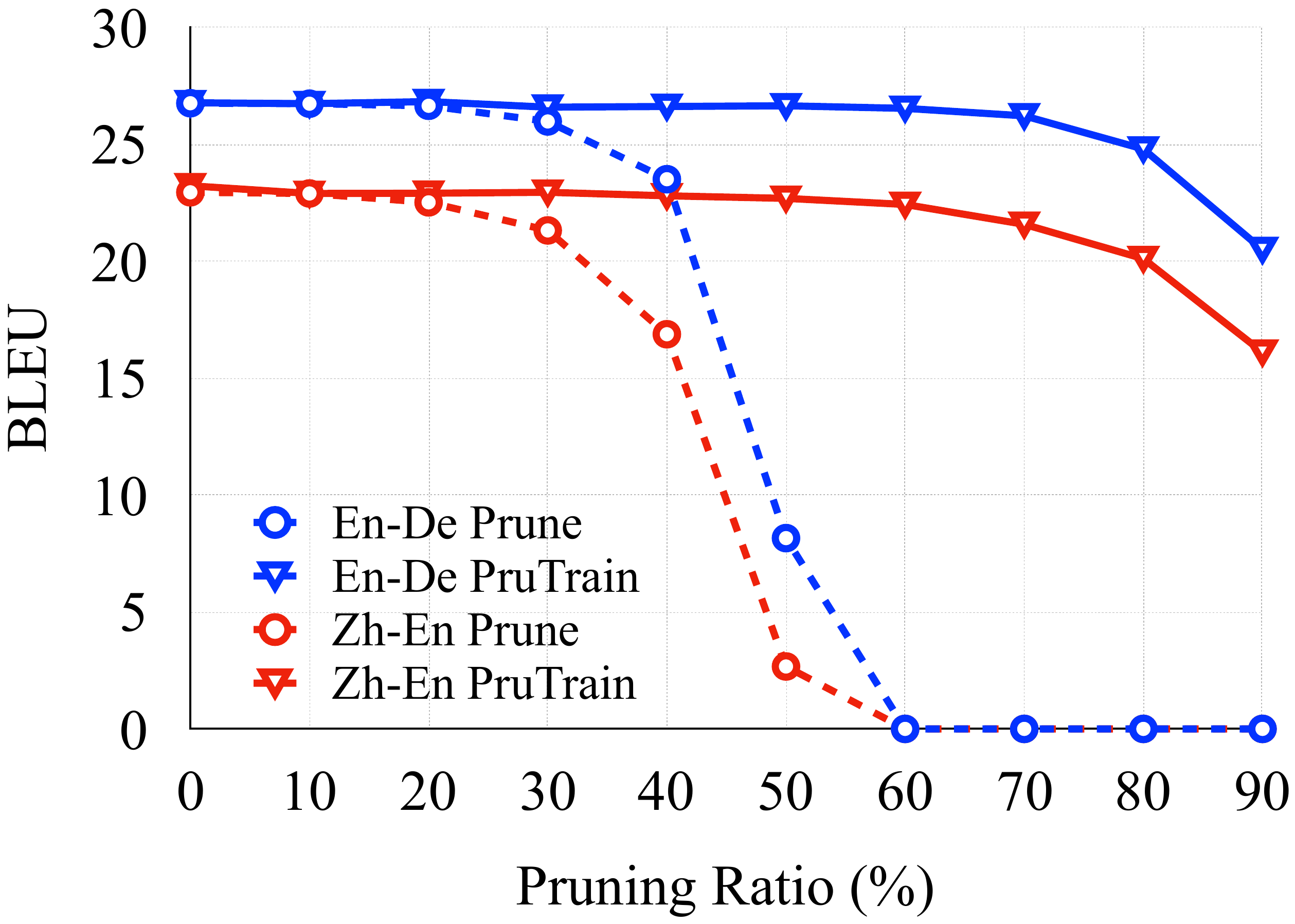}
\caption{Effects of different pruning ratios on Transformer. ``Prune'' denotes directly pruning parameters while ``PruTrain'' indicates adding continuous training after the pruning phase.}
\label{fig-prune-ratios}
\end{figure}

\paragraph{Rejuvenation}
After the pruning and retraining phases, we aim to restore the model capacity by rejuvenating the pruned parameters. This is a common method in optimization to avoid useless computations and further improve performances~\cite{han2017dsd,qiao2018neural}. Thus, we release the sparsity constraint ($a=0, \forall a\in\theta_p$), which inversely recovers the pruned connections, and re-dense the whole networks. The recovered weight connections are then initialized by some strategy (e.g. zero or external). The entire networks are retrained with one order of magnitude lower learning rate since the sparse network is already at a good local optimum. As seen, the rejuvenation method contains three phases: 1) training a baseline model (\textsc{Base}); 2) pruning $\gamma$\% parameters and then retraining remaining ones (PruTrain); 3) restoring pruned parameters and training entire networks (RejTrain).

\begin{table}[t]
\centering
\renewcommand\arraystretch{1.1}
\begin{tabular}{cc|cc|r}
    \multicolumn{2}{c|}{\bf Pruning} & \multicolumn{2}{c|}{\bf Rejuvenation} & \multirow{2}{*}{\bf BLEU} \\
    \cline{1-2} \cline{3-4}
	local & global & zero & external &  \\
	\hline
\checkmark & \texttimes  & \checkmark & \texttimes & 28.12 \\
\texttimes & \checkmark & \checkmark & \texttimes & 28.08 \\
\hdashline
\checkmark & \texttimes & \checkmark & \texttimes & 28.12 \\
\checkmark & \texttimes & \texttimes & \checkmark & 28.14 \\
\end{tabular}
\caption{Effects of different strategies on Transformer on WMT14 En$\Rightarrow$De. ``zero'' denotes using zero as initialization, while ``external'' denotes using corresponding parameters in the baseline model as initialization.}
\label{tab-results-en-de-strategy}
\end{table}

\section{Experiments}
\subsection{Setup}
\paragraph{Data}
We conduct experiments on English$\Rightarrow$ German (En$\Rightarrow$De), Chinese$\Rightarrow$English (Zh$\Rightarrow$En), German$\Rightarrow$English (De$\Rightarrow$En) and English$\Rightarrow$French (En$\Rightarrow$Fr) translation tasks. For En$\Rightarrow$De task, we use WMT14 corpus which contains 4 million sentence pairs. The Zh$\Rightarrow$En task is conducted on WMT17 corpus, consisting of 21 million sentence pairs. We follow \newcite{dou2018exploiting} to select the development and test sets. Furthermore, we evaluate low-resourced translation on IWSLT14 De$\Rightarrow$En and IWSLT17 En$\Rightarrow$Fr corpora. We preprocess our data using byte-pair encoding~\cite{sennrich2016subword} with 40K merge operations for En$\Rightarrow$De, 32K for Zh$\Rightarrow$En, and 10K for De$\Rightarrow$En and En$\Rightarrow$Fr, and keep all tokens in the vocabulary.
We use 4-gram BLEU score~\cite{papineni2002bleu} as the evaluation metric and sign-test~\cite{koehn2004statistical} for statistical significance. 

\paragraph{Models}
We implement our approach on top of three popular architectures, namely Transformer~\citep{vaswani2017attention}, RNNSearch~\citep{luong2015effective} and LightConv~\citep{wu2018pay} with the open-source toolkit -- fairseq~\citep{ott2019fairseq}. For Transformer, we investigate \textit{big}, \textit{base} and \textit{small} settings. About RNNSearch and LightConv, we employ corresponding configurations in fairseq. The implementation is detailed in Appendix \S A.1. All baseline models are trained for 100K updates using Adam optimizer~\cite{kingma2014adam}. Based on the baselines, the proposed pruning and rejuvenation methods are trained with additional 100K updates (i.e. 50K for each one). To rule out the circumstance that more training steps may bring improvements, we also conduct continuous training (ConTrain) as strong baselines and they employ the same training steps as our approach. 

\begin{table}[t]
\centering
\renewcommand\arraystretch{1.1}
\begin{tabular}{c|l|r|lc}
	\bf \# & \bf Model & \bf \# Para. & \bf BLEU & \bf $\Delta$ \\
	\hline
    1 & \textsc{Base}  &  108.6M  & 27.54 & -- \\
    2 &  ~~ + ConTrain  & 108.6M  & 27.74 & +0.20\\\hdashline
    3 & ~~ + RejTrain & 108.6M& $\text{28.12}^\Uparrow$ & +0.58\\
	4 & ~~~~ + RejTrain & 108.6M & $\text{28.33}^\Uparrow$ & +0.79 \\
	\hline
    5 & \textsc{Big} & 305.3M  & 28.55 & -- \\
    6 &  ~~ + ConTrain & 305.3M & 28.81 & +0.26 \\\hdashline
    7 & ~~ + RejTrain & 305.3M & $\text{29.12}^\Uparrow$ & +0.57\\
\end{tabular}
\caption{Translation quality of Transformer model on WMT14 En$\Rightarrow$De. ``\# Para.'' denotes the trainable parameter size of each model. ``+'' denotes appending new features to the above row. ``${\uparrow}/{\Uparrow}$'' indicates statistical significance (${p}<{0.05/0.01}$) over the baseline.}
\label{tab-results-en-de}
\end{table}

\subsection{Results of Pruning}
To study the effect of sparsity, we investigate the effects of different pruning ratios on Transformer \textit{base} models. Experiments are conducted on WMT14 En$\Rightarrow$De and WMT17 Zh$\Rightarrow$En tasks. As shown in Figure~\ref{fig-prune-ratios}, over 20\% of parameters can be directly pruned without degrading the translation performance. When adding a simple continuous training phase after pruning, we are able to prune 50\% with no performance loss. Compared with findings in \citet{see2016compression}, Transformer is less over-parameterized than RNN-based NMT models (20\% vs. 40\% and 50\% vs. 80\%). This provides the evidence that different NMT models are over-parameterized to a different extent. Accordingly, we set the pruning threshold of 50\% as a default in the following experiments (i.e. Tables 1$-$4).

\subsection{Results of Rejuvenation}

\paragraph{Ablation Study}

As shown in Table~\ref{tab-results-en-de-strategy}, we systematically compare different pruning and rejuvenation strategies on the translation task. 
As seen, the local pruning strategy performs better than the global one, especially with the rejuvenation counterpart (28.12 vs. 28.08 BLEU). However, \citet{see2016compression} found that the global pruning outperforms the local one without considering rejuvenation factors. Regarding the rejuvenation strategy, zero and external initialization perform similarly in terms of BLEU score. Therefore, we use local pruning and zero initialization strategies for the rest of the experiments (i.e. Tables 2$-$4).

\begin{table}[t]
\centering
\renewcommand\arraystretch{1.1}
\begin{tabular}{c|l|lc}
	\bf Data & \bf Model  & \bf BLEU & \bf $\Delta$ \\
	\hline
	\multirow{3}{*}{\shortstack{Zh-En \\(21M)}}& 
    \textsc{Base}  & 24.18 & --\\
    & ~~ + ConTrain  & 24.35 & +0.17 \\ \cdashline{2-4}
    & ~~ + RejTrain & $\text{24.60}^\uparrow$ & +0.42 \\\hline
	\multirow{3}{*}{\shortstack{De-En \\(0.16M)}}&
    \textsc{Small}  & 30.50 & --\\
    & ~~ + ConTrain  & 30.50 & +0.00 \\ \cdashline{2-4}
    & ~~ + RejTrain & $\text{30.87}^\Uparrow$ & +0.37 \\\hline
    \multirow{3}{*}{\shortstack{En-Fr \\(0.22M)}}&
    \textsc{Small}  & 38.43 & --\\
    & ~~ + ConTrain  & 38.43 & +0.00  \\ \cdashline{2-4}
    & ~~ + RejTrain & $\text{38.97}^\Uparrow$ & +0.54 \\
\end{tabular}
\caption{Translation quality of Transformer model on different datasets varied in language pair and size.}
\label{tab-results-other-data}
\end{table}

\paragraph{Main Results}
We evaluate the rejuvenation approach on the Transformer using En$\Rightarrow$De dataset. As shown in Table~\ref{tab-results-en-de} (Rows 1$-$4), our model~(RejTrain) outperforms the baseline model and continuous training method~(ConTrain) by +0.58 and +0.38 BLEU points, respectively. In addition, iterative rejuvenation can incrementally improve the baseline model up to 28.33 BLEU points (+0.79 and +0.59 over \textsc{Base} and ConTrain). The results clearly demonstrate the effectiveness of rejuvenating redundant parameters for NMT models.

To verify the robustness, we evaluate different model sizes. As shown in Table~\ref{tab-results-en-de} (Rows 5$-$7), the Transformer \textsc{Big} model performs better than the \textit{base} with an increase of 196.7M parameters. Surprisingly, the performance can be further improved by +0.57 BLEU points by our method. As seen, the continuous training can only slightly gain +0.2 BLEU over \textsc{Big}, and RejTrain outperforms the strong baseline. This confirms that the rejuvenation method can consistently improve NMT models by alleviating the over-parameterization issue.

\begin{table}[t]
\centering
\renewcommand\arraystretch{1.1}
\begin{tabular}{c|l|r|lc}
	\bf \# & \bf Model & \bf Para. & \bf BLEU & \bf $\Delta$ \\
	\hline
    1 & Transformer  &  108.6M  & 27.54 & -- \\
    2 &  ~~ + ConTrain  & 108.6M  & 27.74 & +0.20\\\hdashline
    3 & ~~ + RejTrain & 108.6M & $\text{28.12}^\Uparrow$ & +0.58\\
	\hline
    4 & RNNSearch & 197.0M  & 22.98 & -- \\
    5 &  ~~ + ConTrain & 197.0M  & 22.98 & +0.00 \\\hdashline
    6 & ~~ + RejTrain & 197.0M & \text{23.30} & +0.32\\
    \hline
    7 & LightConv & 304.2M  & 28.01 & -- \\
    8 &  ~~ + ConTrain & 304.2M  & 28.32 & +0.31 \\\hdashline
    9 & ~~ + RejTrain & 304.2M & $\text{28.52}^\Uparrow$ & +0.51\\
\end{tabular}
\caption{Translation quality of different NMT models on WMT14 En$\Rightarrow$De.}
\label{tab-results-arch}
\end{table}

\paragraph{Different Datasets}

Table~\ref{tab-results-other-data} shows results on three datasets: Zh$\Rightarrow$En, De$\Rightarrow$En and En$\Rightarrow$Fr, covering large-scale and small-scale training data (i.e. 21M, 0.16M and 0.22M). Trained with large-scale data (Zh$\Rightarrow$En), the continuous training achieves +0.17 BLEU point over the baseline while the rejuvenation approach obtains +0.42 improvement. For low-resource translation (De$\Rightarrow$En and En$\Rightarrow$Fr), ConTrain can not further improve the performance since it is easy to get stuck in a local optimum. However, RejTrain can jump out of local optimum with improved performances (+0.37 and +0.54 over De$\Rightarrow$En and En$\Rightarrow$Fr baselines, respectively). Compared with continuous training, the proposed method significantly and incrementally improves the translation quality in all cases. This again demonstrates the effectiveness of our method across different datasets varied in aspects of language and size.

\paragraph{Different Model Architectures}

As shown in Table~\ref{tab-results-arch}, we conduct the experiments on WMT14 En$\Rightarrow$De translation task with RNNSearch, LightConv, Transformer models. Our approach achieves consistent and significant improvements over the baseline and ConTrain models across three architectures. For RNNSearch, continuous training cannot further improve the performance while our model achieves better performance (+0.32 over ConTrain). Furthermore, LightConv works better than the Transformer \textsc{Base} model since it has 3$\times$ more parameters. However, RejTrain still outperforms the ConTrain model and achieves 28.52 BLEU scores. This demonstrates the effectiveness and universality of our approach.

\section{Analysis}
To better understand the effectiveness of the proposed method, the analyses are carried out in two ways: representation visualization and linguistic probing. Furthermore, we study the translation outputs in terms of adequacy and fluency.

\begin{figure}[t]
\centering
\includegraphics[width=0.48\textwidth]{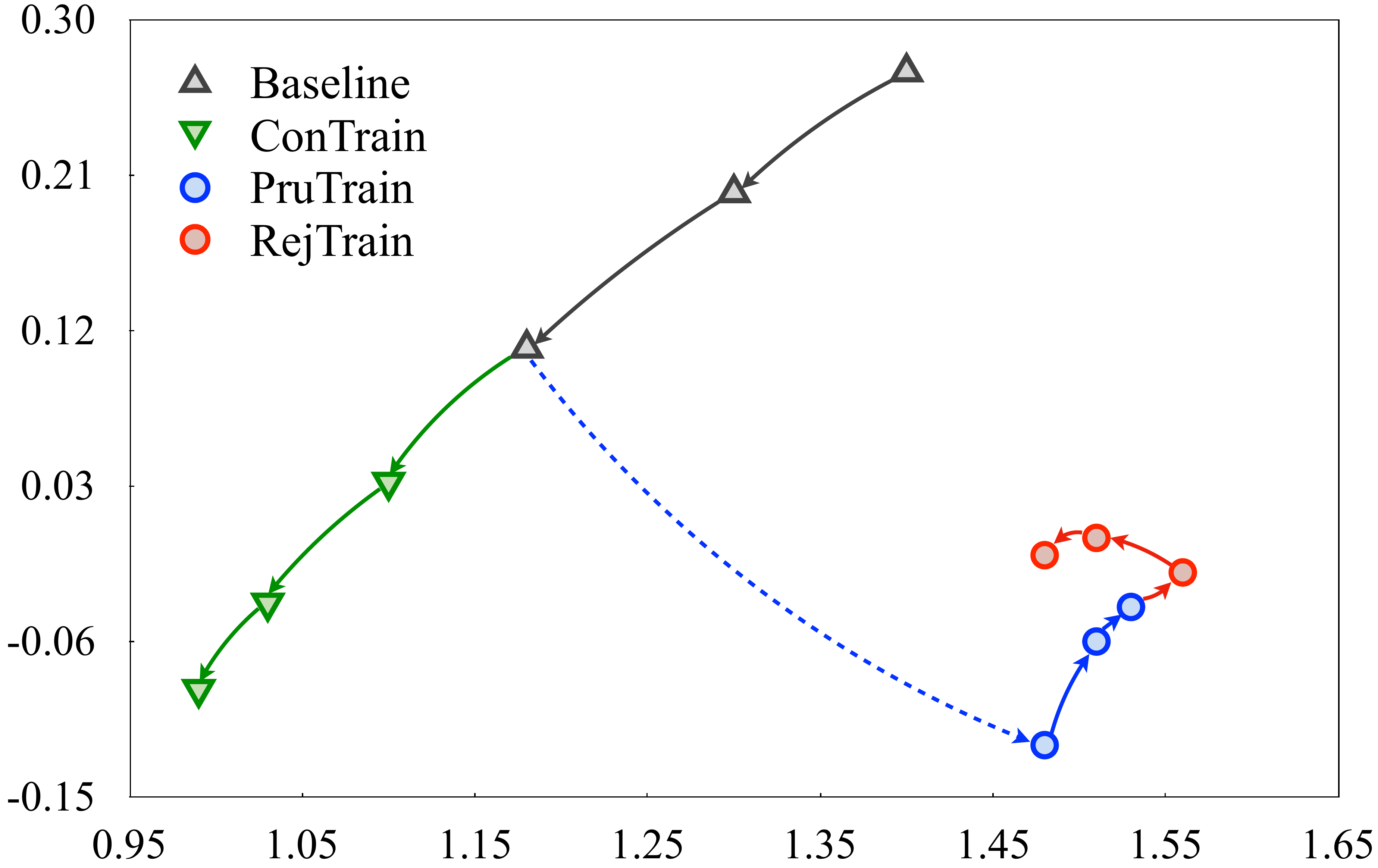}
\caption{Visualization of encoder representations in different training phases. For each phase, we select sequentially three models. The solid arrow represents the changes in each phase. The dotted arrow represents the changes from the baseline to the pruning phase.}
\label{fig-repr-dist}
\end{figure}

\paragraph{Escaping from Local Optimum} 

To study how our method help models to escape from local optimum, we analyze the change of source representations during different training phases. The analysis is conducted on the Transformer \textsc{Base} model and En$\Rightarrow$De. Following \citet{zeng2018multi}, we feed source sentences in the development set into a checkpoint and output an element-wise averaged vector from representations of the last encoder layer. With the dimension-reduction technique of TruncatedSVD~\cite{du2017power}, we can plot the dimensionally reduced values in Figure~\ref{fig-repr-dist}.
Among the training phases (i.e. Baseline, ConTrain, PruTrain, RejTrain), we select checkpoints at which interval training updates are equal. As seen, within each phase, the representations change smoothly in direction and quantity. The continuous training still transforms the representations in the same direction as the baseline phase (i.e. grey vs. green lines). However, the pruning training dramatically changes the representations (i.e. blue vs. grey lines). Finally, the rejuvenation training jumps to a different place compared with the ConTrain (i.e. red vs. green lines). This demonstrates that our method can efficiently change the direction of optimization, thus providing more chances for the model to escape from the local optimum.

\paragraph{Linguistic Insights}

\begin{figure}[t]
\centering
\includegraphics[width=0.45\textwidth]{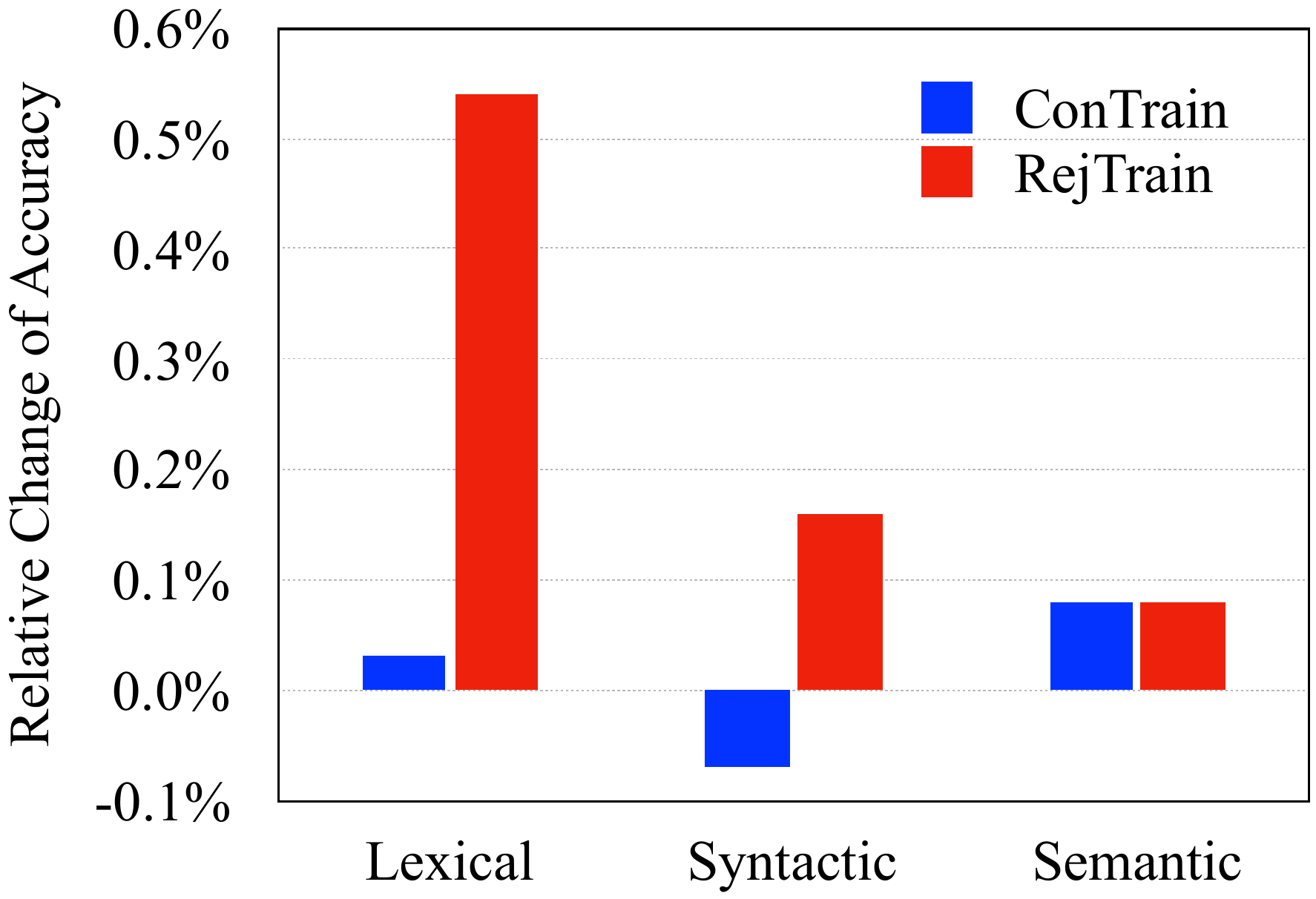}
\caption{Relative change of performance on linguistic probing tasks compared with the baseline. ``Lexical'', ``Syntactic'', ``Semantic'' denote the averaged accuracy over corresponding tasks in each category.}
\label{tab:probing}
\end{figure}

We follow \citet{conneau2018you} to conduct the linguistic probing task, which aims to measure the linguistic knowledge embedded in the encoder representations learned by the model. Specifically, it contains 10 classification subtasks with 3 linguistic categories, including lexical, syntactic and semantic ones. We average the predicted accuracies of subtasks in the same category and calculate the relative changes of ConTrain and RejTrain over the baseline model. The analysis is conducted on the Transformer \textsc{Base} model and En$\Rightarrow$De translation task. As shown in Figure~\ref{tab:probing}, the RejTrain model performs better on the lower level of linguistic subtasks, especially on lexical ones (i.e. 0.6\%). The details are listed in Appendix \S A.2.
The improvements are significant compared with those in \citet{wang2019self}. We hypothesize that better capturing lexical knowledge can improve the adequacy and fluency of translation, which is verified in next part.

\paragraph{Adequacy and Fluency}
\begin{CJK}{UTF8}{gbsn}
Table~\ref{table:example} shows an example randomly selected from the test set in the Zh$\Rightarrow$En task. As seen, incorporating the rejuvenation approach into NMT can generate more fluent translation with higher adequacy. For instance, the Chinese word ``奥运会'' is under-translated by the baseline model, while the RejTrain model can correctly translate it into ``olympics''. Besides, the nominal modifier ``21岁的'' is mistranslated into a simple number by the baseline while RejTrain can fix the error.
This confirms that the rejuvenation improves the adequacy of translation by enhancing the ability to understand the lexical information.
\end{CJK}
To better evaluate the fluency of our models, we calculate the perplexity on the WMT14 En-De test set. 
As shown in Table~\ref{tab-results-ppl-en-de}, the RejTrain model can achieve lower perplexity than baseline and Contrain models (5.08 vs. 5.14/5.15). An interesting finding is that PruTrain increases the perplexity, which may harm the fluency of translation outputs (5.25 vs. 5.14).
This demonstrates that our rejuvenation approach improves the fluency of translation.

\begin{CJK}{UTF8}{gbsn}
\begin{table}[t]
\centering
\renewcommand\arraystretch{1.1}
\begin{tabular}{r|p{5.1cm}} 
\hline
Input & 2000 年 悉尼奥运会 , 已经 21 岁 的 刘璇 已经 来到 了 运动员 生涯 的 末期 。\\
\hdashline
Reference & \textcolor{red}{at the 2000 sydney olympic games} , the already \textcolor{blue}{21-year-old} liu xuan came to the end of his athlete career . \\
\hline
Baseline & \textcolor{red}{in sydney in 2000} , liu xuan , now \textcolor{blue}{21} , has reached the end of his career as an athlete . \\ 
\hdashline
RejTrain & \textcolor{red}{at the 2000 sydney olympics} , the \textcolor{blue}{21-year-old} liu xuan has reached the end of his career . \\ 
\hline
\end{tabular}
\caption{Example of Zh$\Rightarrow$En translation. Phrases colored in \textcolor{red}{red} and \textcolor{blue}{blue} respectively denote adequacy and fluency problems in baseline but fixed by rejuvenation.}
\label{table:example}
\end{table}
\end{CJK}

\begin{table}[t]
\centering
\renewcommand\arraystretch{1.1}
\begin{tabular}{c|l|l}
	\bf \# & \bf Model & \bf PPL  \\
	\hline
    1 & \textsc{Base}   & 5.14 \\
    2 &  ~~ + ConTrain  & 5.15 \\\hdashline
    3 & ~~ + PruTrain & 5.25\\
	4 & ~~ + RejTrain & 5.08 \\
\end{tabular}
\caption{The perplexity of Transformer model on WMT14 En$\Rightarrow$De. ``PruTrain" indicates retraining the remaining parameters after the pruning phase. ``RejTrain" denotes using the rejuvenation approach.}
\label{tab-results-ppl-en-de}
\end{table}

\section{Conclusion}
In this paper, we prove that existing NMT systems are over-parameterized and propose to improve the utilization efficiency of parameters in NMT models by introducing a rejuvenation approach. Empirical results on a variety of language pairs and architectures demonstrate the effectiveness and universality of the presented method. We also analyze the gains from perspectives of learning dynamics and linguistic probing, which give insightful research directions for future work. 

Future directions include continuing the exploration of this research topic for large sequence-to-sequence pre-training models \cite{liu2020multilingual} and multi-domain translation models \cite{wang2019go}. We will employ recent analysis methods to better understand the behaviors of rejuvenated models~\cite{he2019towards,yang2020sublayer}.

\bibliography{emnlp2020}
\bibliographystyle{acl_natbib}

\clearpage
\appendix
\section{Supplemental Material}
\label{sec:supplemental}

\subsection{Experimental Setup}
In the model configuration of Transformer, the \textsc{Base} and \textsc{Big} models differ in the hidden layer size~(512 vs. 1024), filter size~(2048 vs. 4096) and the number of attention heads~(8 vs. 16). The encoder and decoder are composed of a stack of 6 layers. The best model parameters are determined based on the model performance on the development set. All the models are trained on 8 NVIDIA P40 GPUs where each is allocated with a batch size of 4,096 tokens. For IWSLT14 De$\Rightarrow$En and IWSLT17 En$\Rightarrow$Fr tasks, we use the \textsc{small} model, where the encoder and decoder are composed of a stack of 2 layers respectively and which is trained on 1 GPU with a batch size of 4,096 tokens.

\begin{table}[h]
  \centering
  \renewcommand{\arraystretch}{1.1}
  \begin{tabular}{c|c||c|c|c}
\multicolumn{2}{c||}{\bf Model}   & {\textsc{Base}} & ConTrain & RejTrain\\
\hline \hline
\multirow{3}{*}{\rotatebox[origin=c]{90}{{\bf Lexical}}}
&   SeLen   &  91.35\% & 91.40\% & 91.54\%\\
&   WC   &  75.96\% & 75.98\% & 76.85\%\\
\cdashline{2-5}
&   Avg.   &  83.66\% & 83.69\% & 84.20\%\\
\hline
\multirow{4}{*}{\rotatebox[origin=c]{90}{{\bf Syntactic}}}
&   TeDep   &  44.58\% & 44.61\% & 44.67\%\\
&   ToCo   &  76.89\% & 76.71\% & 77.25\%\\
&   BShif   &  72.18\% & 72.11\% & 72.20\%\\
\cdashline{2-5}
&   Avg.   &  64.55\% & 64.48\% & 64.71\%\\
\hline
\multirow{6}{*}{\rotatebox[origin=c]{90}{{\bf Semantic}}}
&   Tense   &  87.61\% & 87.77\% & 88.04\%\\
&   SubN   &  85.25\% & 85.18\% & 85.03\%\\
&   ObjN   &  84.79\% & 84.67\% & 84.57\%\\
&   SoMo   &  53.60\% & 53.30\% & 53.26\%\\
&   CoIn   &  60.85\% & 61.58\% & 61.62\%\\
\cdashline{2-5}
&   Avg.   &  74.42\% & 74.50\% & 74.50\%\\
\end{tabular}
  \caption{Performance on the linguistic probing tasks of evaluating linguistics embedded in the encoder outputs. ``\textsc{Base}'', ``ConTrain'' and ``RejTrain'' respectively denote the baseline model, continuous training and rejuvenation training. ``Avg.'' denotes the average accuracy of each category.} 
  \label{tab:probing-detail}
\end{table}

\subsection{Probing Task}

In order to gain linguistic insights into the learned representations when carrying out the rejuvenation method, we conducted 10 probing tasks~\cite{conneau2018you} to evaluate linguistics knowledge embedded in the final encoding representation learned by the model, as shown in Table~\ref{tab:probing-detail}. From the table, we can see that RejTrain can capture more lexical~(84.20\% vs. 83.66\%) and syntactic~(64.71\% vs. 64.55\%) information.

\end{document}